\documentclass[sigconf]{acmart} 

\usepackage{microtype}
\usepackage{graphicx}
\usepackage{subcaption}
\usepackage{booktabs} %
\usepackage{array}
\usepackage{hyperref}
\usepackage{multirow}
\usepackage{url}
\usepackage{xurl}

\usepackage{amsmath}
\usepackage{mathtools}
\usepackage{amsthm}

\newcommand{\method}{MUTE}
\theoremstyle{plain}

\theoremstyle{definition}

\theoremstyle{remark}

\usepackage[ruled,vlined]{algorithm2e}
\usepackage{pbalance}
\usepackage{algorithmic}

\usepackage{pifont}

\AtBeginDocument{%
  }

\copyrightyear{2026}
\acmYear{2026}
\setcopyright{cc}
\setcctype{by}
\acmConference[MM '26]{Proceedings of the 34th ACM International Conference on Multimedia}{November 10--14, 2026}{Rio de Janeiro, Brazil}
\acmBooktitle{Proceedings of the 34th ACM International Conference on Multimedia (MM '26), November 10--14, 2026, Rio de Janeiro, Brazil}
\acmDOI{10.1145/3767308.3835583}
\acmISBN{979-8-4007-2213-4/2026/11}

\begin{document}

\title{Motion Concept Unlearning in Video Diffusion Models}

\author{Ping Liu}
\orcid{0000-0002-3170-3783}%
\affiliation{%
  \department{Computer Science and Engineering}
  \institution{University of Nevada, Reno}
  \city{Reno}
  \state{NV}
  \country{USA}
}
\email{pino.pingliu@gmail.com}

\author{Chi Zhang}
\authornote{Corresponding author.}
\orcid{0000-0002-5735-4454}
\affiliation{%
  \department{Department of Mathematics}
  \institution{National University of Singapore}
  \city{}
  \country{Singapore}
}
\email{czhang24@nus.edu.sg}

\renewcommand{\shortauthors}{Liu and Zhang}

\begin{abstract}
Text-to-video (T2V) diffusion models can generate realistic depictions of actions such as kicking, stabbing, and shooting, raising safety concerns that motivate targeted concept erasure.
Although concept erasure has been extensively studied for static concepts in text-to-image and T2V models, erasing motion concepts remains largely unexplored.
We present a systematic study of motion concept erasure in video Diffusion Transformers (DiTs).
Through causal interventions, we show that text-conditioning attention carries concept-specific motion information and supports selective intervention, whereas perturbing temporal positional encoding suppresses both target and non-target dynamics.
We further find that directly adapting ESD, a representative weight-level image erasure method, to a video DiT yields modest and uneven motion suppression: reducing its erasure training loss does not by itself remove the concept signal from the difference between the conditional and unconditional predictions, which classifier-free guidance (CFG) then scales at every denoising step.
From these findings, we derive three requirements for motion concept erasure: concept specificity, spatial selectivity, and temporal naturalness.
Each determines one component of \method{} (\textbf{M}otion concept \textbf{U}nlearning in \textbf{T}ext-to-video g\textbf{E}neration): at each denoising step, \method{} extracts a concept direction through token neutralization, derives a spatial gate from the direction's intrinsic structure, and subtracts the resulting correction from the velocity output before CFG is applied.
\method{} is training-free and requires no weight modification.
Experiments on 20 motion concepts show that \method{} outperforms representative prompt-level, weight-level, and inference-time baselines on Wan2.1-T2V, and the same formulation transfers to CogVideoX, supporting its applicability across distinct T2V attention architectures.
\end{abstract}




\maketitle

\section{Introduction}
\label{sec:intro}

Text-to-video diffusion models~\cite{henschel2025streamingt2v_cvpr2025,sun2025t2v_cvpr2025,lin2025stiv_iccv2025,simon2025titan_iccv2025,yuan2025magictime_tpami2025,wan2025wan_arxiv2025} can now generate realistic, temporally coherent videos from textual descriptions.
However, this capability introduces safety risks~\cite{schramowski2023safe_cvpr2023,qu2023unsafe}: these models can readily generate depictions of undesirable actions such as kicking, stabbing, or shooting.
As these models are increasingly deployed in public-facing applications, post-generation safety filters alone are insufficient: they can be bypassed through adversarial prompt rephrasing~\cite{yang2024sneakyprompt,tsai2024ringabell}, motivating concept erasure, which modifies the generative process to suppress the target concept~\cite{gandikota2023esd_iccv2023}.

Concept erasure is well-studied for static concepts such as objects, styles, and identities in text-to-image models~\cite{gandikota2023esd_iccv2023,gandikota2024uce_wacv2024,chavhan2025conceptprune,saha2025side_emnlp2025,nie2025erasing_iclr2025,wu2024doco}, and a few recent works extend it to static concepts in video models~\cite{xu2025videoeraserconcepterasuretexttovideo,ye2025t2vunlearningconcepterasingmethod,huang2025conceptvoid_mathematics2025,xie2026probediagnosingresidualconcept_arxiv2026}.
These works, however, do not address the distinct problem of erasing motion concepts: the dynamic patterns of how things move, such as kicking, punching, or stabbing.
This gap is both practical and technical: existing methods are designed to erase a specific object or an artistic style rather than the way things move.
Unlike static concepts, motion unfolds over time.
Video Diffusion Transformers (DiTs) introduce temporal components absent from image models, including temporal positional encoding.
It is therefore unclear whether motion concepts remain localized in the text-conditioning attention used by image-domain erasure methods, or are instead distributed across these temporal components.

\begin{figure*}[t]
\centering
\includegraphics[width=\textwidth]{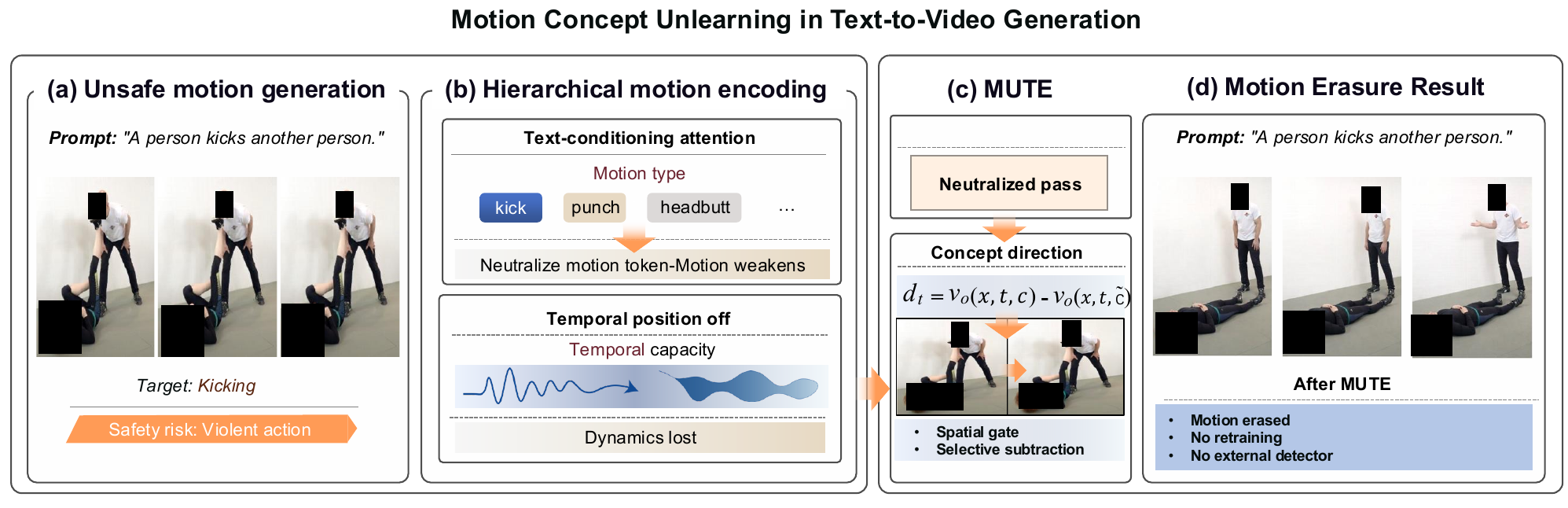}
\caption{Overview of \method{}.
\textbf{(a)}~Text-to-video models generate realistic actions from prompts.
\textbf{(b)}~Probing shows cross-attention carries concept-specific motion and can be selectively intervened on (top), whereas temporal position encoding underlies all dynamics, so suppressing it destroys motion indiscriminately (bottom).
\textbf{(c)}~\method{} extracts the concept direction $d_t$ via token neutralization, derives a spatial gate $M_t$ from $d_t$, and subtracts the concept from the velocity output at each denoising step, with no retraining or external detector.
\textbf{(d)}~The target motion (kicking) is erased while scene composition is preserved under the same prompt. 
}
\label{fig:teaser}
\end{figure*}

Addressing this gap requires understanding where motion information is encoded in video DiTs.
Through causal intervention experiments on Wan2.1-T2V (Section~\ref{sec:probing}), we find that text-conditioning attention, instantiated as cross-attention in this model, carries concept-specific motion information and supports selective intervention.
Temporal positional encoding, by contrast, supports temporal dynamics globally: perturbing it suppresses both target and non-target dynamics (Figure~\ref{fig:teaser}).
Of the two components we probe, text-conditioning attention is therefore the more suitable target for selective erasure.
However, directly adapting ESD~\cite{gandikota2023esd_iccv2023}, a representative weight-level image erasure method, to this channel yields modest and uneven suppression, consistent with a residual concept signal that classifier-free guidance (CFG) still scales (Section~\ref{sec:hierarchy_and_limits}).
This motivates an output-level approach: an inference-time intervention in velocity space, where the concept contribution is removed at each denoising step before CFG amplification occurs.

These findings lead to three requirements for motion concept erasure: concept specificity, requiring that only the target concept is affected; spatial selectivity, restricting the intervention to motion-relevant regions; and temporal naturalness, preserving non-target dynamics.
We show that these requirements, combined with the probing findings, directly determine each component of our method (Section~\ref{sec:method}).
The resulting method, \method{}, operates at inference time with no weight modification.
At each denoising step, \method{} extracts the concept direction via token neutralization, derives a spatial gate from the direction's intrinsic spatial structure, and subtracts the correction from the velocity output before CFG amplification.

We validate \method{} on 20 motion concepts across two video DiTs: Wan2.1-T2V~\cite{wan2025wan_arxiv2025} with separate cross-attention and CogVideoX~\cite{yang2024cogvideox_arxiv2024} with joint attention.
In both, \method{} operates on the text-conditioning attention: the separate cross-attention layer in Wan2.1-T2V and the joint attention layer in CogVideoX.
With the same formulation and algorithm in both architectures, \method{} suppresses the majority of target motions while preserving scene appearance.
Our work makes three contributions.
\begin{itemize}
    \item We formulate motion concept erasure as a problem distinct from static concept erasure and, through causal interventions, identify text-conditioning attention as a viable channel for selective motion erasure.
    Temporal positional encoding, in contrast, supports dynamics globally and is unsuitable for concept-specific intervention.
    We further find that directly adapting ESD to text-conditioning attention yields only modest and uneven suppression, motivating an output-level approach that removes concept contributions from the velocity prediction at each denoising step.
    \item We derive three requirements from the probing analysis and show that each determines a component of \method{}: token neutralization for motion concept isolation, a self-derived spatial gate for localized correction, and per-step velocity subtraction before CFG amplification.
    \item We validate \method{} on 20 motion concepts: it achieves stronger suppression than negative prompting, UCE, VideoEraser, and ESD on Wan2.1-T2V, while the same formulation and algorithm transfer to CogVideoX using a fixed correction strength for each architecture.
\end{itemize}

\section{Related Work}
\label{sec:related}

\subsection{Concept Erasure in Text-to-Image Models}

Existing concept erasure methods for T2I diffusion models intervene on different internal components, including cross-attention, FFN layers, and broader denoiser parameters~\cite{xie2025erasing_arxiv2025}.
\textcolor{black}{Weight-modification approaches divide into optimization-based methods, which fine-tune cross-attention or backbone parameters to steer the model away from the target concept~\cite{gandikota2023esd_iccv2023,fan2024salun}, and closed-form editing methods on cross-attention layers without gradient updates~\cite{gandikota2024uce_wacv2024,Zhang2026DP_cvpr2026}; hybrid designs with lightweight adapters~\cite{lu2024mace}.}
Pruning-based methods instead remove concept-responsive neurons in FFN or attention layers~\cite{chavhan2025conceptprune,cywinski2025saeuron}.
Recent work has expanded along several axes, including continual erasure under sequential concept arrival~\cite{lee2025continual_iclr2026,han2025ccrt,tu2026massconcepterasurediffusion}, training-free inference-time approaches~\cite{lee2025gloce,kim2026safetyguided_iclr2026}\textcolor{black}{, stronger closed-form formulations~\cite{Zhang2026DP_cvpr2026}}, retain-forget entanglement~\cite{cheng2026machine}, extension to rectified flow architectures~\cite{gao2025eraseanything}\textcolor{black}{, autoregressive image generators~\cite{shakibania2026obliviate_eccv2026}, and single-stream diffusion transformers~\cite{jiang2026z_icml2026}}.
\textcolor{black}{A complementary direction makes erasure context-sensitive, deciding whether a concept should be suppressed based on contextual intent rather than removing it unconditionally~\cite{liu2026abcde_tmlr2026}. Separately, recent work~\cite{liu2025erased_arxiv2025} investigates the permanence of concept erasure, questioning whether erased concepts are truly eliminated or instead persist as latent representations that can be reactivated.}

\subsection{Concept Erasure in Text-to-Video Models}

Concept erasure in video diffusion models is a nascent area, with most existing work targeting static concepts.
Early efforts adapt image-domain techniques to video editing: Liu et al.~\cite{liu2025unlearningconceptstexttovideodiffusion} apply fine-tuning-based unlearning to video diffusion backbones, and T2VUnlearning~\cite{ye2025t2vunlearningconcepterasingmethod} extends this direction with concept-specific loss formulations.
Training-free erasure and safeguarding approaches have also emerged: VideoEraser~\cite{xu2025videoeraserconcepterasuretexttovideo} erases objects and NSFW content at inference time, SAFREE~\cite{yoon2025safree} provides an adaptive safety guard for both image and video generation, and Instant Concept Erasure~\cite{biswas2025itdontinstant} performs modality-agnostic erasure through closed-form spectral unlearning applicable to both T2I and T2V models.
ConceptVoid~\cite{huang2025conceptvoid_mathematics2025} addresses multi-concept removal in video diffusion.
On the safety detection side, ConceptGuard~\cite{ma2025conceptguardproactivesafetytextandimagetovideo} takes a complementary approach through multimodal risk detection for text-and-image-to-video generation.
PROBE~\cite{xie2026probediagnosingresidualconcept_arxiv2026} diagnoses residual concept capacity after erasure, providing evaluation infrastructure for this emerging area.
\textcolor{black}{More recently, CLEAR~\cite{xie2026concept_icml2026} studies where concept erasure should occur, while~\cite{facchiano2026video_iclr2026} suppresses unsafe concepts in video generators through a low-rank refusal vector applied at the weight level.}
\textcolor{black}{Although its target categories include violence, it treats them as broad safety concepts rather than formulating or analyzing the selective erasure of individual motion patterns.}
Overall, existing methods primarily target static concepts such as objects, identities, and styles, or the safety-driven refusal of unsafe content.
The selective erasure of a specific motion concept while preserving the actors and non-target dynamics, together with an analysis of where motion is encoded in video DiTs, remains largely unexplored and poses challenges that we address in this work.

\subsection{Attention Analysis and Control in Diffusion Models}
The growing scale and over-parameterization~\cite{zhang2021distributed} of modern generative models have rendered full-model retraining infeasible, placing fine-tuning at the center of model adaptation and editing~\cite{hu2022lora,zhang2024parameter,zhang2025weight}. Understanding how attention encodes information in diffusion models therefore has been an active area of study.
For U-Net-based models~\cite{rombach2022high,saharia2022photorealistic}, prior work shows that cross-attention carries the category-level semantics of conditioning tokens while self-attention encodes spatial layout~\cite{liu2024towards_cvpr2024,sun2025attentive_aaai2025}.
Generated content can also be attributed to individual conditioning tokens through cross-attention~\cite{tang2023daam_acl2023}, while a parallel line of work manipulates attention maps for training-free generation control~\cite{hertz2022prompt_arxiv2022,chefer2023attend_tog2023,epstein2023diffusion_nips2023}.
For multimodal diffusion transformers with joint text-vision attention, recent work decomposes joint attention into semantic-alignment and consistency components and extracts token-precise attention for concept detection~\cite{shen2025qk_iccv2025,li2025detect_cvpr2025}.
Building on this line, we probe both text-conditioning attention and temporal positional encoding in video DiTs to identify where motion information resides and use the resulting structure to derive a motion-erasure method.

\section{Preliminaries}
\label{sec:prelim}

\subsection{Flow Matching and Velocity Prediction}

\textcolor{black}{Wan2.1-T2V, the model we use for probing and for our main experiments, is trained with the flow matching framework}~\cite{wan2025wan_arxiv2025,lipman2022flow_arxiv2022,liu2022rectified_arxiv2022}, \textcolor{black}{which defines a probability path between the data and noise distributions through a time-dependent interpolation.}
Given a data sample $x_0$ and noise $\epsilon \sim \mathcal{N}(0, I)$, the forward interpolation at time $t \in [0, 1]$ is:
\begin{equation}
    x_t = (1 - t)\, x_0 + t\, \epsilon.
    \label{eq:flow_interp}
\end{equation}
A neural network $v_\theta$ is trained to predict the velocity field that transports $x_t$ along this path, with the training objective:
\begin{equation}
    \mathcal{L} = \mathbb{E}_{x_0, \epsilon, t} \bigl[\| v_\theta(x_t, t, c) - (\epsilon - x_0) \|_2^2 \bigr],
    \label{eq:flow_loss}
\end{equation}
where $c$ is the text condition.
At inference, \textcolor{black}{Wan2.1-T2V~\cite{wan2025wan_arxiv2025} generates samples by solving} the ordinary differential equation $\mathrm{d}x_t = v_\theta(x_t, t, c)\,\mathrm{d}t$ from $t\!=\!1$ to $t\!=\!0$ using a numerical solver.
\textcolor{black}{CogVideoX is instead trained under a standard diffusion forward process $x_t = \alpha_t x_0 + \sigma_t \epsilon$ with the $v$-prediction parameterization~\cite{yang2024cogvideox_arxiv2024,salimans2022progressive_iclr2022}, whose target $v_t = \alpha_t \epsilon - \sigma_t x_0$ is likewise a velocity-type quantity.}
\textcolor{black}{\method{} operates on the model's native per-step velocity output and its response to a change in the text condition; its formulation is therefore independent of which of the two parameterizations is used.}
\textcolor{black}{The concept direction $d_t$ and influence map $I_t$ defined in Section~\ref{sec:method} are both computed from this native velocity output.}

\subsection{Text-to-Video Architectures and Conditioning}

Many recent text-to-video models adopt the Diffusion Transformer (DiT)~\cite{peebles2023scalable_iccv2023} architecture but differ in how they handle text conditioning and positional encoding.
\textcolor{black}{We study two designs that differ along both axes.}
The first design, exemplified by Wan2.1-T2V~\cite{wan2025wan_arxiv2025}, uses a sequence of transformer blocks, each containing self-attention over the full visual token sequence, a separate cross-attention layer where visual tokens attend to umT5-XXL text embeddings~\cite{chung2023unimax_iclr2023}, and a feed-forward network with AdaLN timestep modulation~\cite{peebles2023scalable_iccv2023,wan2025wan_arxiv2025}.
\textcolor{black}{Its self-attention uses factorized three-dimensional rotary positional embeddings (RoPE)~\cite{su2024roformer_neurocomputing2024,wan2025wan_arxiv2025}, with separate frequency components for time, height, and width.}
The second design, exemplified by CogVideoX~\cite{yang2024cogvideox_arxiv2024}, also uses a sequence of transformer blocks but concatenates text and visual tokens into a single sequence processed by one joint attention layer per block, with no separate cross-attention; it encodes text with a T5-XXL text encoder~\cite{raffel2020exploring_jmlr2020}, \textcolor{black}{and the 2B variant used in our experiments uses fixed three-dimensional sinusoidal positional embeddings.}

\textcolor{black}{Despite these differences, both models inject the text condition through an attention layer, which we refer to as the text-conditioning attention (the separate cross-attention layer in Wan2.1-T2V and the joint attention layer in CogVideoX), and both use CFG~\cite{ho2021classifierfree_nipsw2021} during inference:}
\begin{equation}
    v_{\text{cfg}} = v_\theta(x_t, t, \varnothing) + s \cdot \bigl(v_\theta(x_t, t, c) - v_\theta(x_t, t, \varnothing)\bigr),
    \label{eq:cfg}
\end{equation}
where $\varnothing$ is the null condition and $s$ is the guidance scale.

\section{Probing Motion Encoding}
\label{sec:probing}

To determine where motion information resides in video DiTs, we conduct causal intervention experiments on Wan2.1-T2V, targeting two encoding channels: text-conditioning cross-attention and temporal positional encoding.
While prior work has established that cross-attention carries concept semantics in image diffusion models~\cite{hertz2022prompt_arxiv2022,liu2024towards_cvpr2024,tang2023daam_acl2023}, video DiTs introduce temporal positional encoding as an additional channel absent in image models, and it is unclear a priori whether the image-domain assumption transfers to motion concepts or whether motion information is distributed differently across these channels.
To quantify changes in target-motion presence, we use the motion consistency score (MCS), an X-CLIP~\cite{ni2022expanding_eccv2022} measure (extending CLIP~\cite{radford2021learning_icml2021} to video) that is higher when the target motion is present.
We report $\Delta$MCS $= \text{MCS}_\text{baseline} - \text{MCS}_\text{condition}$, where positive values indicate weakened target motion.
The metric and prompt construction are detailed in Section~\ref{sec:exp_setup}.

\subsection{Cross-Attention versus Temporal Positional Encoding}
\label{sec:probing_crossattn}
We begin by examining cross-attention, the primary pathway through which text prompt information is injected into the video generation process.
For each target motion, we construct matched motion/static prompt pairs (e.g., ``A person kicks another person'' vs.\ ``A person stands facing another person'').
We zero the cross-attention context embeddings of the motion-verb tokens across all 30 blocks during inference, suppressing the motion instruction while leaving the rest of the prompt intact.
Table~\ref{tab:encoding_patterns} reports results across six motion concepts.
The effect varies across concepts: push ($+1.53$), bite ($+1.07$), and kick ($+0.72$) show the largest reductions, headbutt ($+0.56$) and slap ($+0.42$) smaller ones, and punch ($-0.10$) essentially no change.
These results indicate that cross-attention carries concept-specific motion information, but the degree of dependence varies: some motions are substantially weakened when their text tokens are neutralized, while others show less response.
This raises the question of whether motion is also supported by other architectural channels.

\begin{table}[t]
\centering
\caption{Effect of zeroing motion-verb token embeddings in cross-attention across all 30 transformer blocks. $\Delta$MCS $= \text{MCS}_\text{baseline} - \text{MCS}_\text{zeroed}$; positive values indicate motion weakened by the intervention.}
\label{tab:encoding_patterns}
\footnotesize
\begin{tabular}{l|cccccc}
\toprule
 & kick & push & bite & punch & slap & headbutt \\
\midrule
$\Delta$MCS $\uparrow$ & +0.72 & +1.53 & +1.07 & $-$0.10 & +0.42 & +0.56 \\
\bottomrule
\end{tabular}
\end{table}

\begin{table}[t]
\centering
\caption{ESD vs.\ \method{} on six motion concepts; positive $\Delta$MCS = motion suppressed.}
\label{tab:esd_vs_mute}
\footnotesize
\begin{tabular}{l|cccccc|c}
\toprule
 & kick & push & bite & punch & slap & headbutt & Mean \\
\midrule
$\Delta$MCS (ESD) & +0.18 & $-$0.05 & +1.64 & $-$0.67 & $-$0.30 & +0.57 & +0.23 \\
$\Delta$MCS (\method{}) & +0.21 & +0.98 & +1.79 & $-$0.08 & +2.83 & +0.20 & \textbf{+0.99} \\
\bottomrule
\end{tabular}
\end{table}

We then probe temporal positional encoding~\cite{su2024roformer_neurocomputing2024,wan2025wan_arxiv2025}.
We scale the temporal RoPE rotation angles by a factor $\kappa$ ($e^{i\theta} \to e^{i\theta \cdot \kappa}$) toward zero, collapsing all frames to a shared temporal position.
At $\kappa\!=\!0$, qualitative inspection across the tested concepts shows that the generated videos become nearly static, losing both the target motion and non-target dynamics such as camera and ambient motion.
This suggests that temporal RoPE supports temporal dynamics globally rather than encoding an individual motion concept, making it unsuitable for selective erasure.

The key question for erasure is therefore not which component affects motion, but which can be manipulated selectively without collapsing temporal dynamics globally.
Of the two channels probed, cross-attention is the more suitable target for selective motion erasure: intervening on motion-token conditioning can weaken individual motions without causing the global temporal collapse observed when temporal RoPE is suppressed.

\subsection{Adapting Weight-Level Erasure to Motion Concepts}
\label{sec:hierarchy_and_limits}

Once cross-attention is identified as a viable selective channel, a natural first strategy is to adapt weight-level cross-attention erasure from text-to-image diffusion models, where this approach has been successful for static concepts.
As a representative and widely used weight-level method, we adapt ESD~\cite{gandikota2023esd_iccv2023} to Wan2.1-T2V by fine-tuning all cross-attention query, key, and value projections with negative guidance loss for 1000 steps across six motion concepts.
Under this fine-tuning budget, all six runs reduce the ESD erasure objective by roughly 97\%, while the resulting suppression is modest and uneven.
Mean $\Delta$MCS is $+0.23$, with 3 of 6 concepts showing positive values: bite reaches $+1.64$, whereas punch and slap fall to $-0.67$ and $-0.30$ (Table~\ref{tab:esd_vs_mute}).
For reference, zeroing the motion-token context embeddings at inference time yields a mean $\Delta$MCS of $+0.70$ on the same six concepts (Table~\ref{tab:encoding_patterns}).
This probe is not a usable erasure method, since it removes the motion instruction from the prompt itself, but the comparison indicates that intervening on this channel at inference time can reach a level of suppression that the weight edit did not.

One possible reading of this gap comes from a simple scaling view of CFG.
Let $\delta_t$ denote the target-concept component of the difference between the conditional and unconditional predictions before weight editing, and let $(1-\rho_t)\,\delta_t$ denote the residual component that remains after editing.
Under CFG (Eq.~\eqref{eq:cfg}), this residual enters the guided prediction as $s\,(1-\rho_t)\,\delta_t$, and its aggregate contribution to the denoising trajectory can be approximated as
\begin{equation}
    R \approx \sum_{t=1}^{T} a_t \, s \, (1-\rho_t) \, \delta_t,
    \label{eq:cfg_residual}
\end{equation}
where $a_t$ denotes the effective update coefficient of the numerical solver at step $t$.
The residual term $(1-\rho_t)\,\delta_t$ is therefore scaled by the guidance weight and propagated along the denoising trajectory.
The reduction in the ESD training objective does not directly determine $\rho_t$, but the suppression we observe in Table~\ref{tab:esd_vs_mute} is consistent with residual motion signals remaining in the difference between the conditional and unconditional predictions.\footnote{We report this as one configuration rather than an upper bound on weight-level erasure: more targeted or higher-cost fine-tuning may reduce the residual further. We instead pursue a training-free route that does not require weight-level fine-tuning.}
This suggests a complementary, state-dependent route: instead of relying on a single weight edit to hold across all latent states, we estimate the target contribution at each denoising step and subtract it from the model output immediately before CFG is applied.

\section{MUTE: Motion Concept Erasure}
\label{sec:method}

\subsection{Problem Formulation}
\label{sec:method_formulation}

Given that directly adapting weight-level erasure yields limited and inconsistent suppression (Section~\ref{sec:hierarchy_and_limits}), we formulate motion erasure as an output-level intervention on the step-wise velocity prediction.
Let $v_\theta(x_t, t, c)$ be the velocity prediction of a video DiT at denoising step $t$, conditioned on text $c$.
Given a target motion concept $\mathcal{C}$ (e.g., ``kicking''), we seek a correction term $\Delta_\mathcal{C}$ such that:
\begin{equation}
    v_\text{erased}(x_t, t, c) = v_\theta(x_t, t, c) - \Delta_\mathcal{C}(x_t, t, c).
    \label{eq:formulation}
\end{equation}
The formulation is intentionally general; the key question is what properties $\Delta_\mathcal{C}$ should satisfy to remove the target motion without degrading the rest of the video.
The analysis in Section~\ref{sec:probing} imposes three requirements on any valid correction term.
\textbf{(R1)}~Concept specificity: $\Delta_\mathcal{C}$ should isolate only the contribution of the target motion concept, so that suppressing ``kick'' removes the kicking action itself without altering unrelated semantic content or non-target actions.
\textbf{(R2)}~Spatial selectivity: $\Delta_\mathcal{C}$ should be concentrated on motion-relevant spatial regions rather than diffusing across the entire frame, preserving scene elements such as the person, objects, and background.
\textbf{(R3)}~Temporal naturalness: $\Delta_\mathcal{C}$ should suppress the target motion while preserving non-target temporal dynamics, including subtle body motion and camera movement, so that the resulting video remains temporally plausible.

\subsection{Concept Direction (R1)}
\label{sec:method_derivation}

R1 requires isolating the target concept's contribution from the velocity output.
Since our probing identifies text-conditioning attention, instantiated as cross-attention in Wan2.1-T2V, as the channel carrying concept-specific motion information, a natural step-wise proxy for the concept's contribution is the change in velocity induced by neutralizing the target token signal while keeping the rest of the condition fixed.
Concretely, after text encoding, we construct a modified condition $\tilde{c}$ by jointly replacing the encoder-output context embeddings at all target-keyword subword positions (e.g., ``kicks'') with the mean embedding over the full fixed-length context sequence, while leaving all remaining positions unchanged.
We refer to this operation as token neutralization.
Because it acts on the context sequence rather than on a particular attention layout, the same operation applies to both separate and joint attention architectures.
The concept direction at step $t$ is then:
\begin{equation}
    d_t = v_\theta(x_t, t, c) - v_\theta(x_t, t, \tilde{c}).
    \label{eq:concept_dir}
\end{equation}
The resulting $d_t$ serves as a step-wise estimate, rather than an exact decomposition, of the target token's contribution to the velocity output (R1).

\subsection{Self-Derived Spatial Gate (R2, R3)}

\begin{figure}[t]
\centering
\includegraphics[width=\columnwidth]{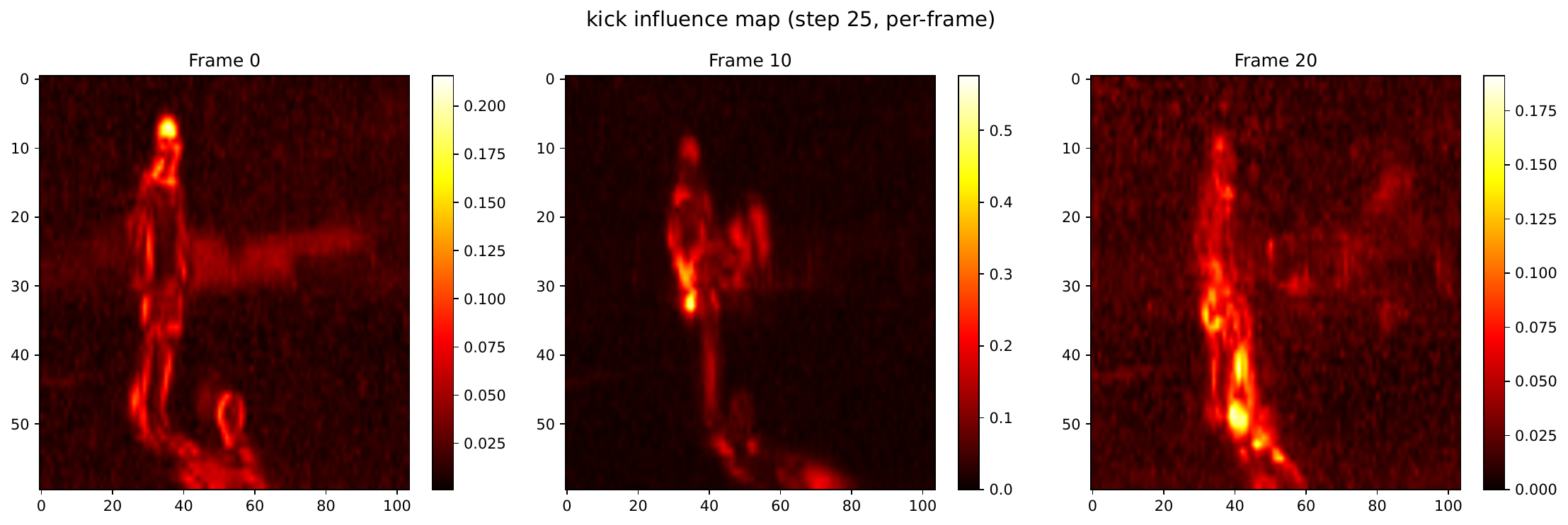}
\caption{Influence map $I_t$ for ``kicking'' at denoising step $t\!=\!25$ across three frames. The influence concentrates on the human body and contacted regions, with much smaller values in the background, providing an intrinsic spatial gate without any external segmentation model (R2).}
\label{fig:influence_map}
\end{figure}

The concept direction $d_t$ satisfies R1, but applying it uniformly ($v_\text{erased} = v_\theta - \alpha \cdot d_t$) would modify all spatial locations including the background, violating R2.
We observe that $d_t$ is in fact spatially concentrated, with a peak-to-mean ratio exceeding 20:1 in the cases we examine: it is large in regions where the target motion occurs and small elsewhere, and its spatial structure adapts to the target concept (Figure~\ref{fig:influence_map}).
This is consistent with the cross-attention mechanism, where the motion token's attention weights are concentrated on the spatial positions depicting the action.
We exploit this by computing the per-position channel-norm as an influence map:
\begin{equation}
    I_t(i,j,k) = \| d_t(:, i, j, k) \|_2.
    \label{eq:influence_raw}
\end{equation}
The normalized influence map
\begin{equation}
    M_t = \frac{I_t}{\max(I_t)},
    \label{eq:influence_norm}
\end{equation}
therefore serves as a self-derived spatial gate: $M_t \approx 1$ in motion-relevant regions (R2) and $M_t \ll 1$ elsewhere, preserving non-target dynamics (R3), without any external segmentation model.\footnote{In practice, we add $\epsilon = 10^{-8}$ to $\max(I_t)$ for numerical stability.}

\subsection{Full Algorithm and Properties}

\begin{algorithm}[t]
\caption{\method{}: Motion Concept Erasure}
\label{alg:mute}
\begin{algorithmic}[1]
\REQUIRE Prompt $c$, target motion keyword, strength $\alpha$, CFG scale $s$
\STATE Identify target token positions in $c$; construct $\tilde{c}$ by mean replacement
\IF{no target token found}
    \STATE \textbf{return} standard generation
\ENDIF
\STATE $x_T \sim \mathcal{N}(0, I)$
\FOR{$t = T, T{-}1, \ldots, 1$}
    \STATE $v_\text{cond} \gets v_\theta(x_t, t, c)$
    \STATE $v_\text{neutral} \gets v_\theta(x_t, t, \tilde{c})$ \hfill \texttt{// R1: concept isolation}
    \STATE $d_t \gets v_\text{cond} - v_\text{neutral}$
    \STATE $I_t(i,j,k) \gets \|d_t(:,i,j,k)\|_2$;\, $M_t \gets I_t / \max(I_t)$ \hfill \texttt{// R2, R3}
    \STATE $v_\text{erased} \gets v_\text{cond} - \alpha \cdot M_t \odot d_t$ \hfill \texttt{// selective erasure}
    \STATE $v_\varnothing \gets v_\theta(x_t, t, \varnothing)$
    \STATE $v_\text{final} \gets v_\varnothing + s \cdot (v_\text{erased} - v_\varnothing)$
    \STATE $x_{t-1} \gets \text{SchedulerStep}(v_\text{final}, x_t, t)$
\ENDFOR
\STATE \textbf{return} $\text{VAE.decode}(x_0)$
\end{algorithmic}
\end{algorithm}

Combining the concept direction with the spatial gate yields the complete correction term:
\begin{equation}
    \Delta_\mathcal{C}(x_t, t, c) = \alpha \cdot M_t \odot d_t,
    \label{eq:delta}
\end{equation}
and the erased velocity:
\begin{equation}
    v_\text{erased} = v_\theta(x_t, t, c) - \alpha \cdot M_t \odot d_t,
    \label{eq:selective_erase}
\end{equation}
which replaces the conditional velocity in standard CFG:
\begin{equation}
    v_\text{final} = v_\varnothing + s \cdot (v_\text{erased} - v_\varnothing).
    \label{eq:cfg_erased}
\end{equation}
Substituting Eq.~\eqref{eq:selective_erase} into Eq.~\eqref{eq:cfg_erased} and comparing with standard CFG in Eq.~\eqref{eq:cfg} yields a compact form:
\begin{equation}
    v_\text{final} = v_\text{cfg} - s \cdot \alpha \cdot M_t \odot d_t,
    \label{eq:mute_compact}
\end{equation}
where $v_\text{cfg}$ is the standard CFG output. Therefore,
\method{}'s output equals the unmodified CFG output minus a scaled concept correction.
This reveals a structural parallel with CFG itself: CFG uses $v_\text{cond} - v_\varnothing$ to amplify all conditioning; \method{} uses $v_\text{cond} - v_\text{neutral}$ to isolate and remove the targeted concept, acting as concept-specific negative guidance with spatial selectivity.
Eq.~\eqref{eq:mute_compact} also provides a quantitative interpretation of $\alpha$: the target concept's contribution in $v_\text{cfg}$ is approximately $s \cdot d_t$ (in regions where $M_t \approx 1$).
If $d_t$ perfectly captured the concept's per-step contribution, $\alpha\!=\!1$ would suffice for complete erasure.
In practice, token neutralization may underestimate the actual concept contribution, as some motion information can be distributed across non-target tokens through contextual interactions in the text encoder and cross-attention layers~\cite{liu2024towards_cvpr2024}, so $\alpha > 1$ compensates for this gap, consistent with the ablation in Table~\ref{tab:ablation_alpha}.
\textcolor{black}{Sweeping $\alpha$ over an order of magnitude in that ablation produces a smooth erasure-fidelity trade-off rather than abrupt degradation, so the single hyperparameter can be set without fine-grained tuning.}

Algorithm~\ref{alg:mute} summarizes \method{}.
It requires neither training nor weight modification, and has a single hyperparameter $\alpha$ controlling erasure strength.
Each denoising step uses three forward passes (conditional, neutralized, unconditional) rather than two, a $\sim$47\% wall-clock overhead.
Unlike weight-level methods such as ESD, which must suppress the concept implicitly through a single fixed weight edit (Section~\ref{sec:hierarchy_and_limits}), \method{} estimates the concept contribution from the current latent state and removes it from the velocity output before CFG is applied at each step.

\section{Experiments}
\label{sec:exp}

\subsection{Setup}
\label{sec:exp_setup}

We evaluate on Wan2.1-T2V-1.3B~\cite{wan2025wan_arxiv2025} (832$\times$480, 81 frames) and CogVideoX-2B~\cite{yang2024cogvideox_arxiv2024} (720$\times$480, 49 frames), targeting 20  motions (kick, punch, slap, push, stab, shoot, trip, bite, choke, headbutt, whip, slash, stomp, tackle, drag, smash, slam, elbow, knee, scratch) with up to 3 prompts per concept\footnote{The full prompt list is provided in the supplementary material.}.
All experiments use $\alpha\!=\!5.0$, spatial gating, mean-replacement neutralization.
We report X-CLIP MCS~\cite{ni2022expanding_eccv2022} (motion consistency score; decrease indicates suppression), LPIPS~\cite{zhang2018unreasonable_cvpr2018} (frame-level perceptual distance), SSIM~\cite{wang2004image_tip2004} (structural similarity), and qualitative inspection.
We compare against representative text-to-image erasure methods originally designed for static concept removal (ESD \cite{gandikota2023esd_iccv2023}, UCE \cite{gandikota2024uce_wacv2024}, and negative prompting~\cite{ho2021classifierfree_nipsw2021}) and text-to-video erasure methods (VideoEraser \cite{xu2025videoeraserconcepterasuretexttovideo}\textcolor{black}{ and T2VUnlearning~\cite{ye2025t2vunlearningconcepterasingmethod}}), which likewise target static concepts rather than motion-specific erasure; baseline implementation details and parameter settings are provided in the supplementary material.

\subsection{Quantitative Results}
\label{sec:exp_main}

\begin{table*}[t]
\centering
\begin{minipage}[t]{0.63\textwidth}
\centering
\captionof{table}{\method{} vs.\ baselines on Wan2.1-T2V-1.3B. $\Delta$MCS $= \text{MCS}_\text{baseline} - \text{MCS}_\text{method}$; positive values indicate stronger motion suppression. LPIPS and SSIM are reported for \method{}.}
\label{tab:main_results_wan}
\footnotesize
\begin{tabular}{l|cccccc|cc}
\toprule
 & \multicolumn{6}{c|}{$\Delta$MCS $\uparrow$} & \multicolumn{2}{c}{\method{} fidelity} \\
Concept & Neg & UCE & ESD & T2VU & VE & \method{} & LPIPS & SSIM \\
\midrule
kick     & +0.26 & +0.31 & $-$0.06 & $-$0.16 & +0.32 & \textbf{+0.75} & .50 & .43 \\
punch    & +0.95 & +0.39 & $-$0.66 & +0.99 & +2.43 & \textbf{+3.81} & .82 & .22 \\
slap     & +0.58 & $-$0.16 & +0.02 & +1.36 & +1.05 & \textbf{+1.87} & .79 & .28 \\
push     & +0.12 & +0.07 & +0.60 & \textbf{+1.54} & +1.37 & +1.42 & .77 & .25 \\
stab     & +1.36 & $-$0.11 & $-$0.93 & +0.23 & \textbf{+1.93} & +1.64 & .82 & .17 \\
shoot    & +1.03 & +0.75 & +0.32 & +2.07 & +1.43 & \textbf{+2.22} & .74 & .24 \\
trip     & $-$0.05 & +0.78 & \textbf{+1.45} & $-$0.41 & +1.31 & +0.91 & .78 & .33 \\
bite     & $-$0.23 & $-$0.58 & \textbf{+2.12} & +0.22 & +0.47 & +0.64 & .60 & .34 \\
choke    & +0.41 & +0.54 & $-$1.57 & +0.45 & \textbf{+2.52} & $-$0.40 & .46 & .54 \\
headbutt & $-$0.69 & +0.09 & +0.99 & \textbf{+2.24} & +0.97 & +0.57 & .51 & .33 \\
whip     & +0.17 & +0.00 & $-$0.45 & +0.30 & +0.95 & \textbf{+1.47} & .79 & .17 \\
slash    & $-$0.44 & $-$0.21 & +1.41 & +0.55 & +1.38 & \textbf{+1.53} & .76 & .23 \\
stomp    & +0.19 & +0.21 & +0.11 & +0.44 & \textbf{+1.00} & +0.79 & .77 & .33 \\
tackle   & +0.05 & $-$0.16 & \textbf{+2.61} & $-$0.98 & +2.15 & +2.14 & .77 & .28 \\
drag     & \textbf{+0.80} & +0.14 & +0.73 & +0.46 & +0.55 & +0.65 & .45 & .48 \\
smash    & +0.14 & +0.33 & +0.70 & +1.04 & +1.33 & \textbf{+1.71} & .79 & .24 \\
slam     & +0.58 & +0.29 & $-$0.12 & +0.42 & +0.55 & \textbf{+1.75} & .84 & .22 \\
elbow    & $-$0.06 & +0.06 & $-$0.12 & $-$0.57 & +0.13 & \textbf{+0.14} & .51 & .40 \\
knee     & +0.11 & +0.14 & $-$0.57 & $-$0.22 & $-$0.21 & \textbf{+0.19} & .58 & .33 \\
scratch  & $-$0.58 & $-$1.02 & $-$0.03 & +0.10 & +0.11 & \textbf{+0.91} & .78 & .29 \\
\midrule
\textbf{Mean} & +0.24 & +0.09 & +0.33 & +0.50 & +1.09 & \textbf{+1.24} & .69 & .31 \\
\bottomrule
\end{tabular}
\end{minipage}
\hfill
\begin{minipage}[t]{0.34\textwidth}
\centering
\captionof{table}{\method{} vs.\ the unmodified base model on CogVideoX-2B. LPIPS and SSIM are reported for \method{}.}
\label{tab:main_results_cog}
\footnotesize
\begin{tabular}{l|cc|cc}
\toprule
 & \multicolumn{2}{c|}{MCS $\downarrow$} & \multicolumn{2}{c}{\method{} fidelity} \\
Concept & Base & \method{} & LPIPS & SSIM \\
\midrule
kick     & $-$0.30 & $-$0.15 & .62 & .47 \\
punch    & 2.78 & 1.65 & .37 & .66 \\
slap     & 0.48 & $-$0.64 & .37 & .67 \\
push     & 0.43 & 0.06 & .52 & .40 \\
stab     & 0.89 & 0.19 & .34 & .64 \\
shoot    & 2.03 & 3.00 & .53 & .57 \\
trip     & $-$0.71 & $-$1.56 & .41 & .47 \\
bite     & 0.25 & 0.44 & .41 & .51 \\
choke    & 2.31 & 1.45 & .39 & .62 \\
headbutt & 1.53 & 1.94 & .44 & .50 \\
whip     & 0.14 & $-$0.17 & .34 & .63 \\
slash    & 0.94 & 0.54 & .40 & .49 \\
stomp    & 0.41 & $-$0.14 & .50 & .53 \\
tackle   & 3.55 & 3.43 & .48 & .48 \\
drag     & $-$0.33 & $-$0.49 & .34 & .67 \\
smash    & 0.63 & 1.13 & .54 & .40 \\
slam     & 1.04 & $-$0.11 & .40 & .57 \\
elbow    & 0.98 & 0.67 & .51 & .50 \\
knee     & 0.13 & 0.19 & .59 & .43 \\
scratch  & 0.88 & 0.81 & .53 & .53 \\
\midrule
\textbf{Mean} & 0.90 & 0.61 & .45 & .54 \\
\bottomrule
\end{tabular}
\end{minipage}
\end{table*}

\begin{figure*}[t]
\centering
\includegraphics[width=1.0\textwidth]{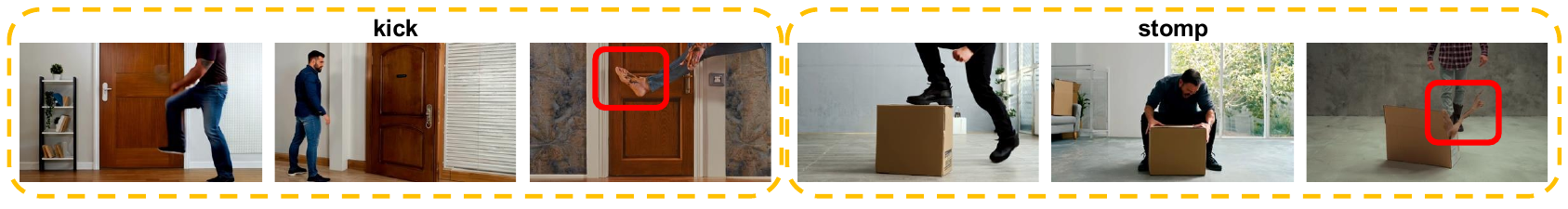}
\caption{Visual artifacts in VideoEraser output.
Each group shows three frames: baseline (L), \method{} (M), and VideoEraser (R).
\textbf{Kick}: shoes on the kicking foot vanish across frames, replaced by a bare foot (R1 violation: appearance removed alongside motion).
\textbf{Stomp}: a phantom hand appears on the cardboard box surface (R2 violation: correction leaks into non-target spatial regions).
\method{} avoids both artifacts through concept-specific direction extraction and spatial gating.}
\label{fig:ve_artifacts}
\end{figure*}

Tables~\ref{tab:main_results_wan} and~\ref{tab:main_results_cog} present per-concept results on both architectures.
On Wan2.1-T2V (Table~\ref{tab:main_results_wan}), \method{} achieves the strongest mean suppression among all methods (mean $\Delta$MCS $+1.24$), substantially outperforming negative prompting~\cite{ho2021classifierfree_nipsw2021} ($+0.24$), UCE~\cite{gandikota2024uce_wacv2024} ($+0.09$), weight-level ESD ($+0.33$)\textcolor{black}{, and the video-specific T2VUnlearning~\cite{ye2025t2vunlearningconcepterasingmethod} ($+0.50$)}.
The gap over these baselines is consistent with the scaling view in Section~\ref{sec:hierarchy_and_limits}: all operate upstream of or at the CFG level, where any residual concept signal is still scaled by the guidance weight.
On CogVideoX (Table~\ref{tab:main_results_cog}), mean MCS drops from 0.90 to 0.61.
The smaller reduction is consistent with its joint attention design: because text and visual tokens share a single attention layer, neutralizing target tokens has a more diffuse effect on the velocity field.
 
The closest competitor is VideoEraser~\cite{xu2025videoeraserconcepterasuretexttovideo}, originally proposed for static concept erasure, which achieves mean $\Delta$MCS $+1.09$ at comparable inference cost.
\method{} achieves stronger per-concept suppression on the majority of concepts, with the advantage most evident on spatially localized actions such as punch ($+3.81$ vs.\ $+2.43$), slam ($+1.75$ vs.\ $+0.55$), and slap ($+1.87$ vs.\ $+1.05$).
Beyond quantitative metrics, qualitative inspection (Figure~\ref{fig:ve_artifacts}) reveals two types of artifacts in VideoEraser's output that \method{} avoids.
First, phantom objects appear on unrelated surfaces (e.g., stomp), indicating that the correction leaks into non-target spatial regions; \method{}'s spatial gate $M_t$ prevents this by confining the intervention to regions where $d_t$ is large (R2).
Second, appearance details such as shoes vanish within the motion region itself (e.g., kick), suggesting that the correction removes scene appearance alongside the target motion; \method{}'s concept direction $d_t$ avoids this because it captures specifically the velocity change induced by the motion token, leaving appearance information largely intact (R1).

Beyond erasure effectiveness, we examine whether scene fidelity is preserved.
On Wan2.1-T2V, mean LPIPS is 0.69 and mean SSIM is 0.31.
On CogVideoX, the corresponding values indicate a smaller visual change (LPIPS 0.45, SSIM 0.54), alongside the weaker suppression noted above.
Per-concept fidelity varies with the spatial extent of the motion: full-body actions such as tackle and slam produce higher LPIPS, while localized actions such as drag and kick yield lower LPIPS, consistent with the spatial gate concentrating corrections on motion-relevant regions.

\begin{figure*}[t]
\centering
\includegraphics[width=0.9\textwidth]{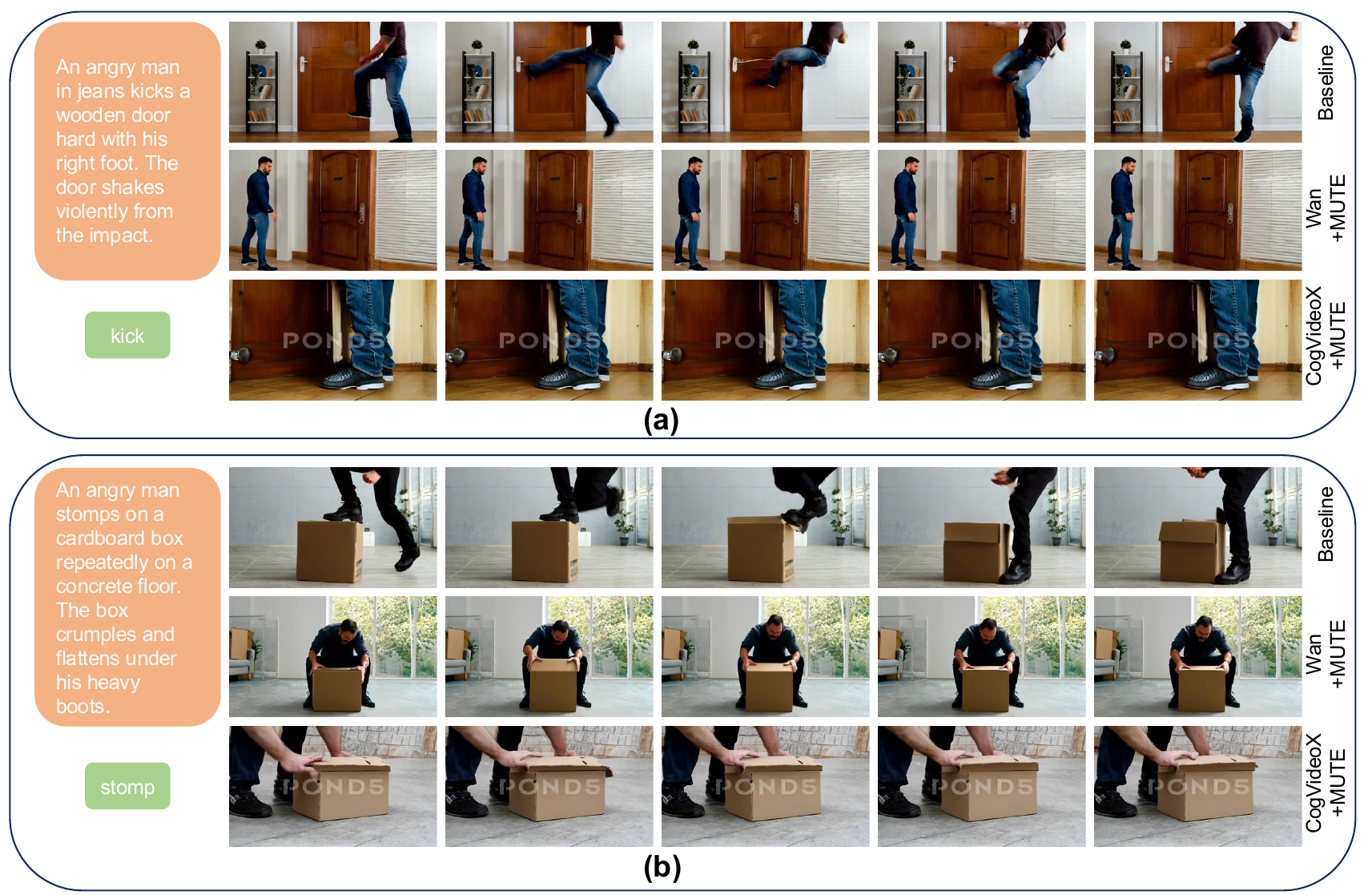}
\caption{Qualitative results for two motion concepts.
Each group shows five evenly-spaced frames from the baseline video (top), Wan2.1-T2V + \method{} (middle), and CogVideoX + \method{} (bottom), all generated from the same prompt.
\textbf{(a)}~Kick: the kicking motion is suppressed while the person, door, and room are preserved.
\textbf{(b)}~Stomp: the stomping action is removed while the person and cardboard box remain intact.
}
\label{fig:qualitative}
\end{figure*}

\begin{figure*}[t]
\centering
\includegraphics[width=0.9\textwidth]{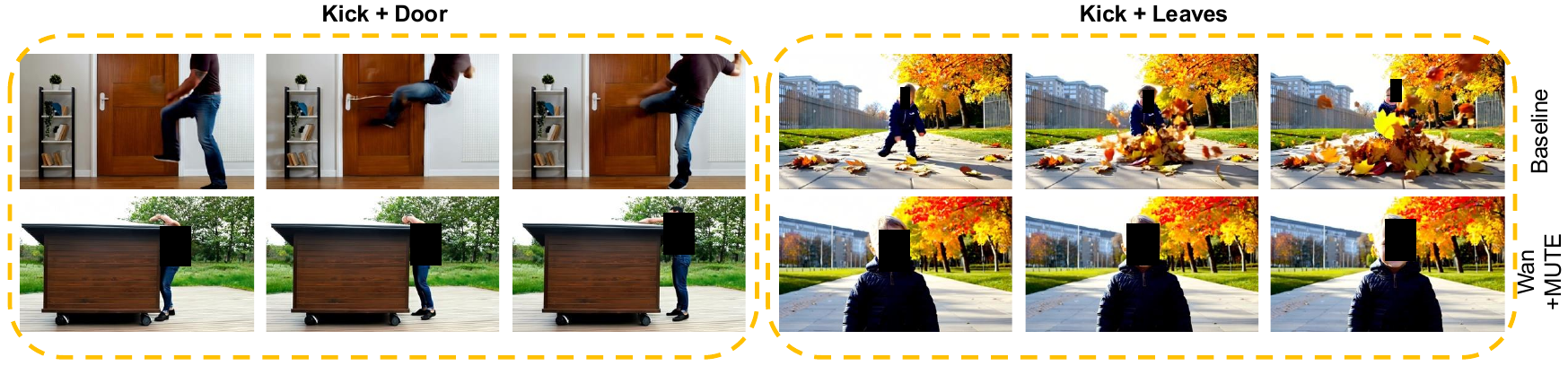}
\caption{Joint motion-object erasure on Wan2.1-T2V.
Each group shows three frames from the baseline (top) and \method{} with joint erasure (bottom).
\textbf{Kick + Door}: the kicking action and the wooden door are both removed; the erased video shows the person standing calmly in an outdoor scene.
\textbf{Kick + Leaves}: the kicking motion and the scattered leaves are both suppressed; the child stands still on the sidewalk with autumn trees preserved in the background.
Although all depicted individuals are AI-generated and do not correspond to real people, we apply facial blurring as a precaution.}
\label{fig:joint_erasure}
\end{figure*}

\subsection{Qualitative Results}
\label{sec:exp_qualitative}

Figure~\ref{fig:qualitative} presents qualitative comparisons for two representative concepts.
For kick, the baseline video (generated by the unmodified model from the same prompt)  shows a person performing a clear kicking motion toward a door; after applying \method{}, Wan2.1-T2V generates the same person standing calmly in front of the door with no kicking action, while the scene (door, room interior, clothing) is preserved.
CogVideoX similarly suppresses the kicking motion, though with a different camera angle consistent with its independent generation process.
For stomp, the baseline depicts a person jumping onto and crushing a cardboard box; \method{} on Wan2.1-T2V replaces this with the person bending over the intact box, removing the stomping while retaining the scene elements (box, floor, background).
In both cases, the spatial gate confines the correction to the motion-relevant regions: the person's legs and contact area for kick, and the person's feet and the box surface for stomp.
Background regions remain visually unchanged, consistent with the spatial concentration of the influence map (Figure~\ref{fig:influence_map}).
Importantly, the erased videos retain natural temporal dynamics: the person continues to exhibit subtle body movement and the camera maintains its original trajectory, indicating that \method{} suppresses the target motion without freezing the scene.

\subsection{Joint Motion-Object Erasure}
\label{sec:exp_multi_concept}
 
\method{} naturally extends to joint erasure of both a motion and an object (e.g., ``kicking'' and ``door'') by computing an independent concept direction and spatial gate for each target type and subtracting both gated corrections from $v_\text{cond}$.
This requires one additional forward pass per step (four total).
Figure~\ref{fig:joint_erasure} illustrates joint erasure on two prompts targeting ``kick'' + ``door'' and ``kick'' + ``leaves''.
In both cases, \method{} removes the kicking action and the associated object simultaneously, with the two influence maps localizing to complementary spatial regions (actor body vs.\ target object).

\subsection{Ablation Studies}
\label{sec:ablation}

We ablate the design choices of \method{} on five representative concepts (kick, punch, slap, bite, tackle) using Wan2.1-T2V.
We first examine the effect of the erasure strength hyperparameter $\alpha$.
Table~\ref{tab:ablation_alpha} sweeps $\alpha$ from 1.0 to 10.0 with spatial gating enabled.
All values successfully suppress motion, but the erasure-fidelity tradeoff is clear: $\Delta$MCS rises from $+1.46$ at $\alpha\!=\!1.0$ to $+1.84$ at $\alpha\!=\!5.0$ and then fluctuates without a consistent trend ($+1.74$ at $\alpha\!=\!7.0$, $+1.91$ at $\alpha\!=\!10.0$), while LPIPS and SSIM degrade monotonically across the whole range.
Beyond $\alpha\!=\!5.0$, larger values therefore bring no reliable gain in suppression at a steady cost in fidelity.
We select $\alpha\!=\!5.0$ as the default.

\begin{table}[t]
\centering
\caption{Ablation on erasure strength $\alpha$. Beyond $\alpha\!=\!5.0$, $\Delta$MCS fluctuates while fidelity continues to degrade. Bold marks the default setting.}
\label{tab:ablation_alpha}
\footnotesize
\begin{tabular}{l|cccccc}
\toprule
$\alpha$ & 1.0 & 2.0 & 3.0 & \textbf{5.0} & 7.0 & 10.0 \\
\midrule
$\Delta$MCS $\uparrow$ & 1.46 & 1.48 & 1.67 & \textbf{1.84} & 1.74 & 1.91 \\
LPIPS $\downarrow$ & 0.604 & 0.650 & 0.671 & \textbf{0.697} & 0.702 & 0.718 \\
SSIM $\uparrow$ & 0.386 & 0.352 & 0.331 & \textbf{0.308} & 0.303 & 0.290 \\
\bottomrule
\end{tabular}
\end{table}

\begin{table}[t]
\centering
\caption{Design component ablation (5 concepts $\times$ 3 prompts). Each row removes or replaces one component. Bold marks the default configuration rather than the best value in each column.}
\label{tab:ablation_design}
\footnotesize
\begin{tabular}{l|c|cc}
\toprule
Configuration & $\Delta$MCS & LPIPS$\downarrow$ & SSIM$\uparrow$ \\
\midrule
\textbf{\method{} (default)} & 1.84 & 0.697 & 0.308 \\
\quad w/ random direction     & 1.19 & 0.503 & 0.508 \\
\quad $\alpha\!=\!1.0$, no gate & 1.53 & 0.647 & 0.351 \\
\quad w/o spatial gate        & 1.97 & 0.694 & 0.310 \\
\bottomrule
\end{tabular}
\end{table}

To isolate the contribution of each component, Table~\ref{tab:ablation_design} presents a systematic ablation.
The most critical component is the concept direction itself (R1): replacing it with a random direction of equal magnitude reduces erasure by 35\%, confirming that the extracted direction carries concept-specific information not replaceable by arbitrary perturbation.
The spatial gate leaves the quantitative metrics largely unchanged ($\Delta$MCS 1.97 vs.\ 1.84, LPIPS 0.694 vs.\ 0.697): its benefit is qualitative, as removing it introduces background artifacts because the subtraction is then applied uniformly across all spatial locations (R2, R3).
Setting $\alpha\!=\!1.0$ without spatial gating, which is equivalent to directly outputting $v_\text{neutral}$, weakens erasure by 17\%, consistent with the quantitative analysis that token neutralization underestimates concept contribution.

\subsection{Human Evaluation}
\label{subsec:human_eval}
We conduct a human study against VideoEraser~\cite{xu2025videoeraserconcepterasuretexttovideo}, the strongest of the compared baselines on the automated metrics (Table~\ref{tab:main_results_wan}), over the 20 motion concepts with one randomly selected prompt each.
Both methods use the same Wan2.1-T2V-1.3B backbone, generation settings, and random seed, so any difference reflects the erasure mechanism rather than backbone or sampling variation.
Seventeen volunteers view the unmodified generation alongside the two erased videos, with method identities hidden, and answer two questions per item: which video better suppresses the target action (\textbf{Motion Suppression}), and which looks more realistic, natural, and artifact-free (\textbf{Visual Quality}).
\method{} receives 57.9\% of votes for motion suppression against 12.1\% for VideoEraser (30.0\% similar), and 82.4\% against 7.6\% for visual quality (10.0\% similar).

\begin{table}[t]
\centering
\caption{Human evaluation results comparing \method{} and VideoEraser (VE) across 20 motion concepts (17 evaluators, 340 pairwise judgments per metric).}
\label{tab:human_eval}
\footnotesize
\begin{tabular}{lccc}
\toprule
Metric & \method{} (\%) & Tie (\%) & VE (\%) \\
\midrule
Motion Suppression & 57.9 & 30.0 & 12.1 \\
Visual Quality & 82.4 & 10.0 & 7.6 \\
\bottomrule
\end{tabular}
\end{table}

\section{Conclusion}
\label{sec:conclusion}

We presented \method{}, a training-free method for selectively erasing motion concepts from text-to-video diffusion models.
Our probing on Wan2.1-T2V shows that cross-attention supports concept-selective intervention, whereas perturbing temporal positional encoding suppresses both target and non-target dynamics.
We further find that a direct adaptation of ESD yields only modest and uneven suppression in this setting.
Guided by these findings, \method{} extracts a concept direction through token neutralization, derives a spatial gate from the direction's intrinsic structure, and subtracts the resulting correction at each denoising step before CFG is applied.
Across 20 motion concepts, \method{} achieves mean $\Delta$MCS $+1.24$ on Wan2.1-T2V, compared with $+1.09$ for the strongest baseline, VideoEraser.
The same default \method{} configuration also transfers to CogVideoX, demonstrating that the approach is not specific to a single T2V attention design.

\clearpage
\bibliographystyle{plain}
\bibliography{main}

\end{document}